\documentclass{article}
\usepackage{spconf}

\usepackage{caption}
\usepackage{subcaption}
\usepackage{cite}
\usepackage{amsmath,amssymb,amsfonts}
\usepackage{algorithmic}
\usepackage{graphicx}
\usepackage{textcomp}
\usepackage[table,xcdraw]{xcolor}
\usepackage{url}
\usepackage{multirow}
\usepackage{booktabs}
\usepackage{balance}

\begin{document}
\ninept 
\title{Privacy-Aware Activity Classification from First Person Office Videos}
\makeatletter
\def\@name{\emph{Partho Ghosh$^{1}$, Md. Abrar Istiak$^{1}$, Nayeeb Rashid$^{1}$, Ahsan Habib Akash$^{1}$, Ridwan Abrar$^{1}$}  
\\ \emph{ Ankan Ghosh Dastider$^{1}$, Asif Shahriyar Sushmit$^{2}$,  and Taufiq Hasan$^{2}$} \vspace{5mm}}
\makeatother

\address{$^{1}$Department of Electrical and Electronic Engineering (EEE)\\ 
$^{2}$mHealth Research Group, Department of Biomedical Engineering (BME)\\
Bangladesh University of Engineering and Technology (BUET), Dhaka - 1205, Bangladesh.}

\maketitle
\begin{abstract}
In the advent of wearable body-cameras, human activity classification from First-Person Videos (FPV) has become a topic of increasing importance for various applications, including in life-logging, law-enforcement, sports, workplace, and healthcare. One of the challenging aspects of FPV is its exposure to potentially sensitive objects within the user's field of view. In this work, we developed a privacy-aware activity classification system focusing on office videos. We utilized a Mask-RCNN with an Inception-ResNet hybrid as a feature extractor for detecting, and then blurring out sensitive objects (e.g., digital screens, human face, paper) from the videos. For activity classification, we incorporate an ensemble of Recurrent Neural Networks (RNNs) with ResNet, ResNext, and DenseNet based feature extractors. The proposed system was trained and evaluated on the FPV office video dataset that includes 18-classes made available through the IEEE Video and Image Processing (VIP) Cup 2019 competition. On the original unprotected FPVs, the proposed activity classifier ensemble reached an accuracy of $85.078\%$ with precision, recall, and F1 scores of $0.88$, $0.85$ \& $0.86$, respectively. On privacy protected videos, the performances were slightly degraded, with accuracy, precision, recall, and F1 scores at $73.68\%$, $0.79$, $0.75$, and $0.74$, respectively. The presented system won the $3$rd prize in the IEEE VIP Cup 2019 competition.
\end{abstract}


\section{Introduction}
As body-worn cameras are becoming more ubiquitous, automatic logging of human activity from FPVs has become topics of increasing interest. At present, FPV or PoV (point-of-view) cameras are mostly utilized by athletes, motor drivers, and police officers \cite{ryoo2013first, chen2019semi}. Generally, it is not feasible to manually analyze large amounts of such videos. Automatically processing FPVs can thus be beneficial in several major applications, including activity logging, law enforcement, search and rescue missions, inspections, home-based rehabilitation, and wildlife observation \cite{tadesse2018visual}. As the Augmented Reality (AR) glasses become mainstream, the availability of FPV data is expected to increase, while also raising associated concerns for security and privacy of the users \cite{roesner2014security}. 

\begin{figure}[t]
    \centering
    \includegraphics[width=\linewidth]{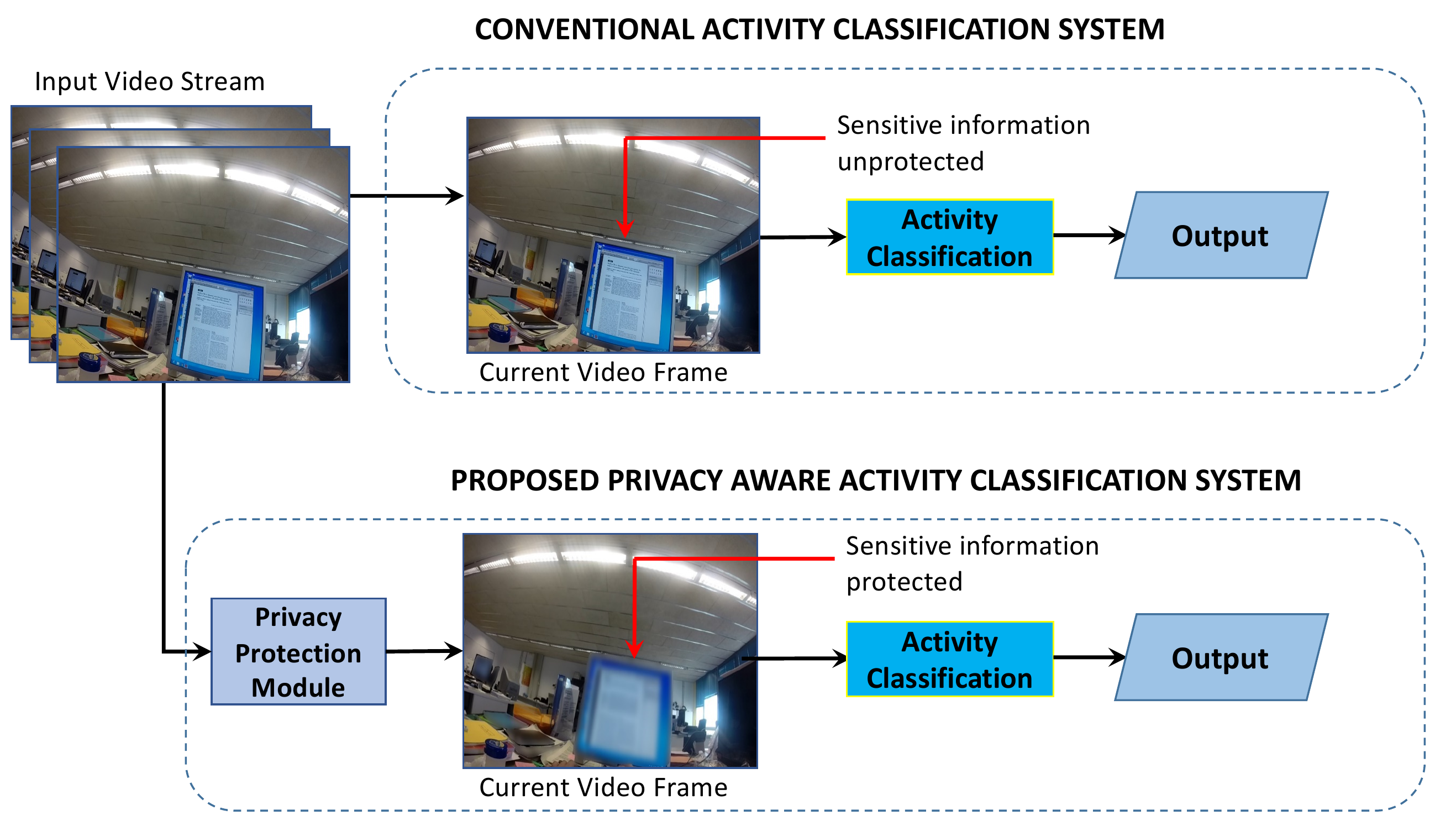}
    \caption{Conventional vs. proposed privacy-aware activity classification systems}
\end{figure}

Automatic activity classification from videos has been undertaken using two major approaches in the past. Traditional methods generally extract hand-engineered features from the videos to train a machine learning algorithm for classification. Frequently employed features include average pooling (AP), robust motion features (RMF), and pooled appearance features (PAF) \cite{abebe2018first,sun2018optical}. In recent times, deep neural networks are being used to learn the features, and subsequently classify the activities. 
One of the effective methods involves using a Convolutional Neural Network (ConvNet) with Long Short-term Memory (LSTM) \cite{hochreiter1997long} units as back-ends \cite{donahue2015long, kahani2019correlation}. In this approach, the ConvNet learns the front-end image features, whereas the LSTMs recognize the temporal features within the videos that are relevant for human activities. Another well-known method found in literature consists of using a 3D ConvNet to recognize the actions \cite{ji20123d}. 3D ConvNets are functionally similar to 2D ConvNets, except that they incorporate Spatio-temporal filters \cite{carreira2017quo}, and thus do not specifically require the LSTM layers. However, none of the previous work in the area of activity classification using FPVs has addressed the issue of privacy.


In this work, we develop a privacy-aware activity classification framework for office videos. In most cases, the personal information contained within FPV videos does not carry useful features for detecting the user's activity. Accordingly, we use a deep learning model to identify the sensitive regions of the video and make them unintelligible (e.g., blurred). Next, these privacy protected videos are used to train deep learning models for activity recognition using the FPV dataset of office activities made available for the IEEE VIP Cup 2019 \cite{abebe2018first}. Performance comparison between activity classification from original vs. privacy protected videos are performed to demonstrate the effectiveness of the proposed framework.

\section{Dataset}
\begin{figure}[bt]
\centering
\includegraphics[width=\linewidth]{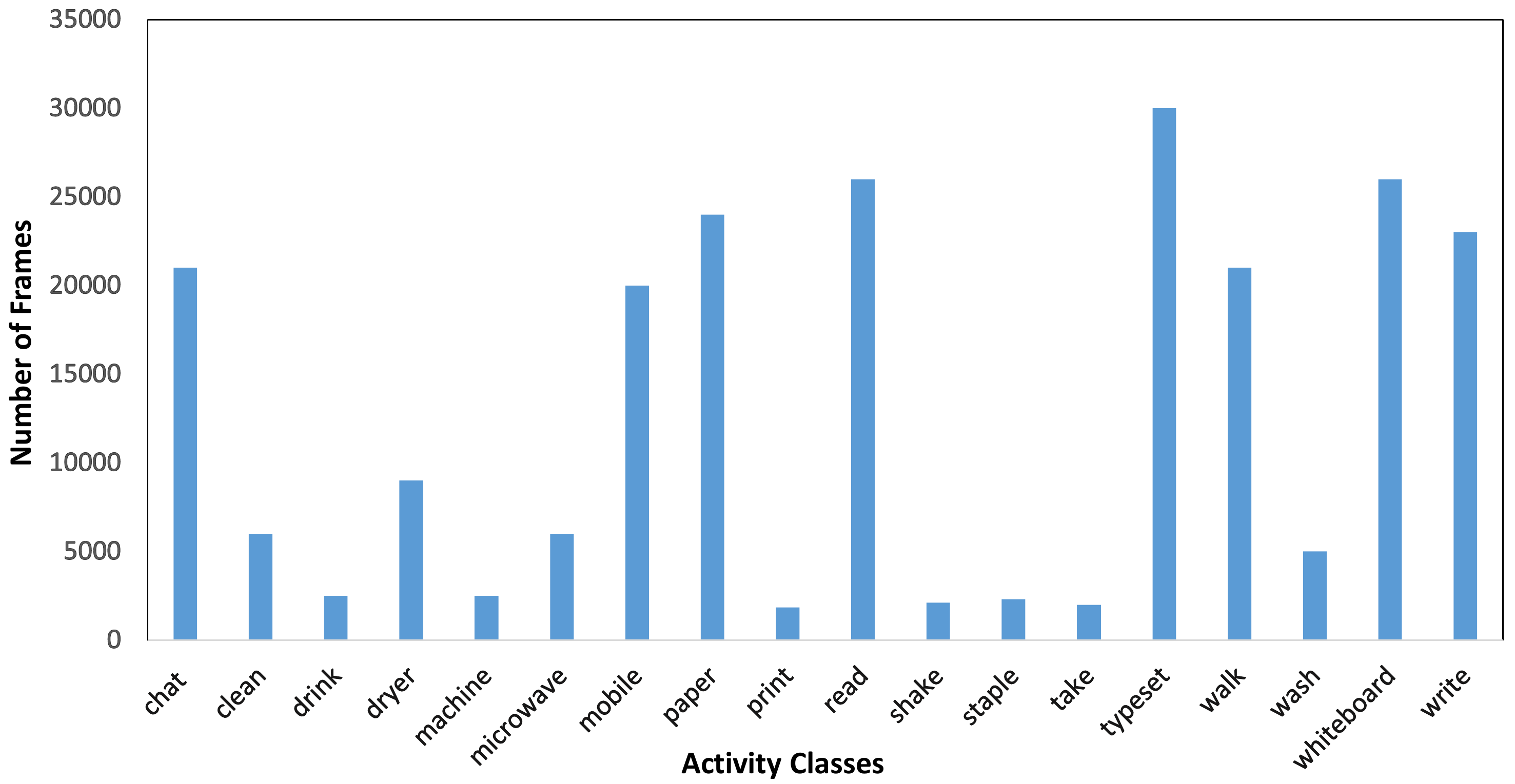}
\caption{Bar plot showing the number of video files available for each activity class. The uneven distribution of data in different classes illustrates the issue of class imbalance.}
\label{class_piechart}
\end{figure}

In this study, we use the FPV office videos provided through the IEEE VIP Cup 2019 competition. We refer to these videos as the FPV-O dataset for the remainder of this paper. The dataset was collected using a chest-mounted GoPro Hero3+ Camera with $1280\times760$ pixels resolution and a frame-rate of 30 fps \cite{abebe2018first}. Human activities can be broadly categorized as (i) human to human activity, (ii) human to object activity and (iii) ambulatory activity. The FPV-O dataset contains a total of $18$ activity in classes that include all of these broad categories. The dataset and distribution of the 18 classes are summarized in Fig. \ref{class_piechart}. As evident from the figure,  the class imbalance problem is a major issue in the dataset that we need to address.

During the competition, the dataset was released in two phases, including $400$ and $932$ videos, respectively. We found label noise in $32$ videos through manual annotation, and these are excluded. The remaining videos from the two phases are used to prepare our training and test set containing $873$ and $364$ videos, respectively.

To train the privacy-aware system, we prepared three different sub-datasets for training and testing our deep learning models. We refer them to as the  (i) \emph{original}, (ii) \emph{blurred}, and (iii) \emph{mixed} sub-datasets. The \emph{original} set contains the original videos from the FPV-O dataset, \emph{blurred} set contains privacy protected videos, and the \emph{mixed} subset contains videos from both subsets (i) and (ii) in an equal amount. The process of generating privacy protected videos in the \emph{blurred}-set are described in the following sections.

\begin{table}[tbh]
\centering
\caption{Training/test splits in the sub-datasets used}
\label{subdata}
\begin{tabular}{|c|c|c|}
\hline
\textbf{Sub-datasets} & \textbf{\# videos in training} & \textbf{\# videos in test} \\ 
\hline
\hline
\emph{Original} & 873 original & 364 original \\ \hline
\emph{Blurred} & 873 blurred & 364 blurred \\ \hline
\emph{Mixed} & \begin{tabular}[c]{@{}c@{}}873 org. + 873 blurred \\(1746 total)\end{tabular} & \begin{tabular}[c]{@{}c@{}}364 org. + 364 blurred\\(728 total)\end{tabular} \\ \hline
\end{tabular}
\end{table}


\section{Proposed Architecture}
\subsection{Pre-processing}
Analyzing the histogram of the average brightness of the video frames (mean value of each image pixel), we observed that some videos have a significantly higher brightness compared to others (e.g., data provided in the ``Oxford" folder). Thus, we applied gamma correction \cite{huang2012efficient}\cite{marcu2006dynamic} to the images to normalize this effect. Next, we resize the video frames and normalize them to fit them in the particular deep learning pre-trained networks, Wide ResNet  ~\cite{zagoruyko2016wide}, ResNext~\cite{xie2017aggregated} and DenseNet~\cite{huang2017densely}, as described later in Sec. \ref{activity_classification}. 

\subsection{Privacy Protection Module}
In this section, we describe the proposed privacy protection and security enhancement module that addresses the privacy concerns in the FPV-O dataset. The framework consists of two steps: (i) Identifying the sensitive regions from a video frame, and (ii) Making these parts unintelligible by blurring \cite{xu2015affect, szegedy2017inception}.

\subsubsection{Identification of sensitive regions}
First, we manually screened the dataset for objects that are prone to privacy violation of the users or may contain sensitive information that can lead to a security breach. In the data, we found a total of 7 objects that may contain sensitive information: (i) digital screen, (ii) laptop, (iii) mobile, (iv) book, (v) person, (vi) keyboard, (vii) toilet/urinal. We used the Mask R-CNN \cite{he2017mask} network to identify these objects in our video frames \cite{lim2011transfer}.

\begin{figure}[t]
    \centering
    \begin{subfigure}[h]{0.5\linewidth}
        \centering
        \includegraphics[width=\linewidth]{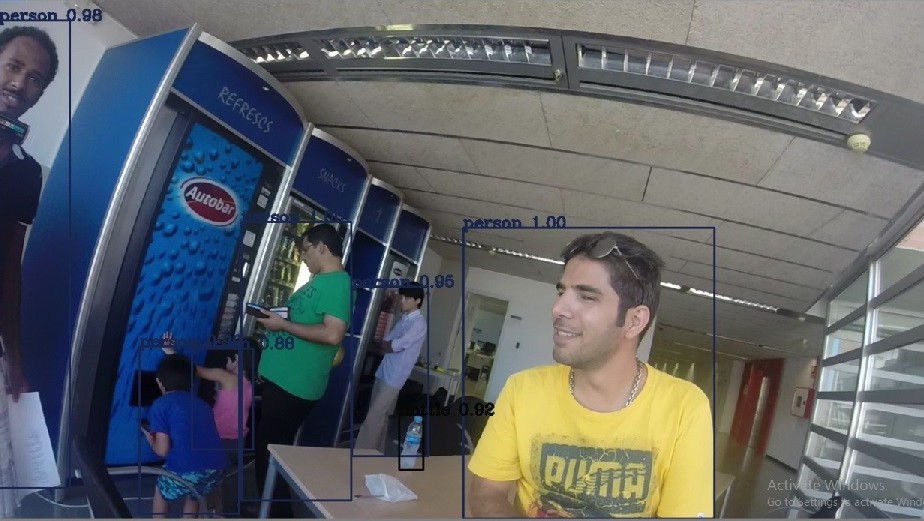}
        \caption{Object localization}
    \end{subfigure}%
    ~ 
    \begin{subfigure}[h]{0.5\linewidth}
        \centering
        \includegraphics[width=\linewidth]{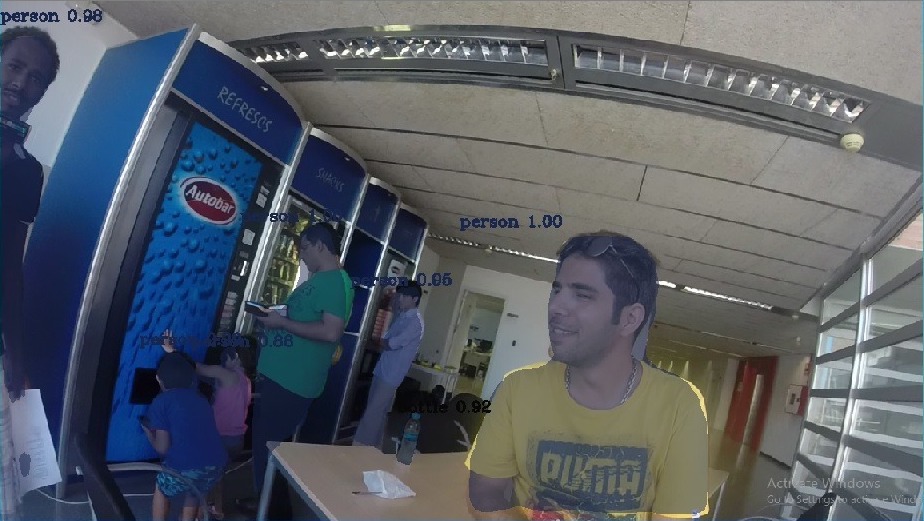}
        \caption{Instance segmentation}
    \end{subfigure}
    \caption{Example image frames illustrating object localization and instance segmentation for a video from obtained from the class ``Chat" of the FPV-O dataset.}
    \label{generated_mask}
\end{figure}

We utilize the Mask R-CNN approach as it performs instance segmentation \cite{romera2016recurrent} to produce a mask for the sensitive objects that may not be of uniform shape (e.g., a person). Traditional methods that generate a bounding-box \cite{long2017accurate,cinbis2016weakly} are not suitable for our application since it will result in the blurring of a larger box-shaped region as compared to the actual sensitive object. This may degrade the activity classifier performance. Examples of object localization (bounding-box) and instance segmentation for an image frame of the activity class ``Chat" is shown in Fig. \ref{generated_mask} (a) and (b), respectively.


\begin{table}[b]
\centering
\caption{Performance of different feature extractor based Mask R-CNN models on the COCO dataset for object segmentation}
\label{mAP}
\begin{tabular}{ll}
\hline
Model Name       & mAP  \\ 
\hline
Mask R-CNN ResNet101 Atrous \cite{rosenfeld2018elephant}                  &33\\
Mask R-CNN Inception V2                      &25 \\
Mask R-CNN ResNet50 Atrous                   &29  \\
Mask R-CNN Inception ResNet V2 Atrous \cite{rosenfeld2018elephant}     &\textbf{36} \\
 \hline
\end{tabular}
\end{table}

Different feature extractor networks can be utilized with a Mask R-CNN model, including InceptionV2 \cite{szegedy2017inception}, Resnet50 \cite{he2016deep}, Resnet101 \cite{he2016deep} and InceptionResNetV2 \cite{szegedy2017inception} with Atrous convolution. The performance of these models for object segmentation, have been compared on the Common Objects in Context (COCO) dataset \cite{lin2014microsoft} in with respect to the Mean Average Precision (mAP) metric \cite{kuznetsova2018open, huang2017speed}. These results are available online in the Tensorflow Github repository\footnote{\url{https://github.com/tensorflow}}. According to these mAP scores reproduced in Table \ref{mAP}, InceptionResNetV2 with Atrous convolution model provides the best performance on the COCO dataset. Thus we select this model for our sensitive object segmentation. 
Since the FPV-O dataset does not contain any object mask labels, it was not possible to objectively evaluate the performance of the object segmentation module on this data. However, we validated the performance of the mask detection model based on visual observation on a sub-set of FPV-O frames. 

\begin{figure}[t]
    \centering
    \includegraphics[trim={20 100 120 20},clip, width=\linewidth]{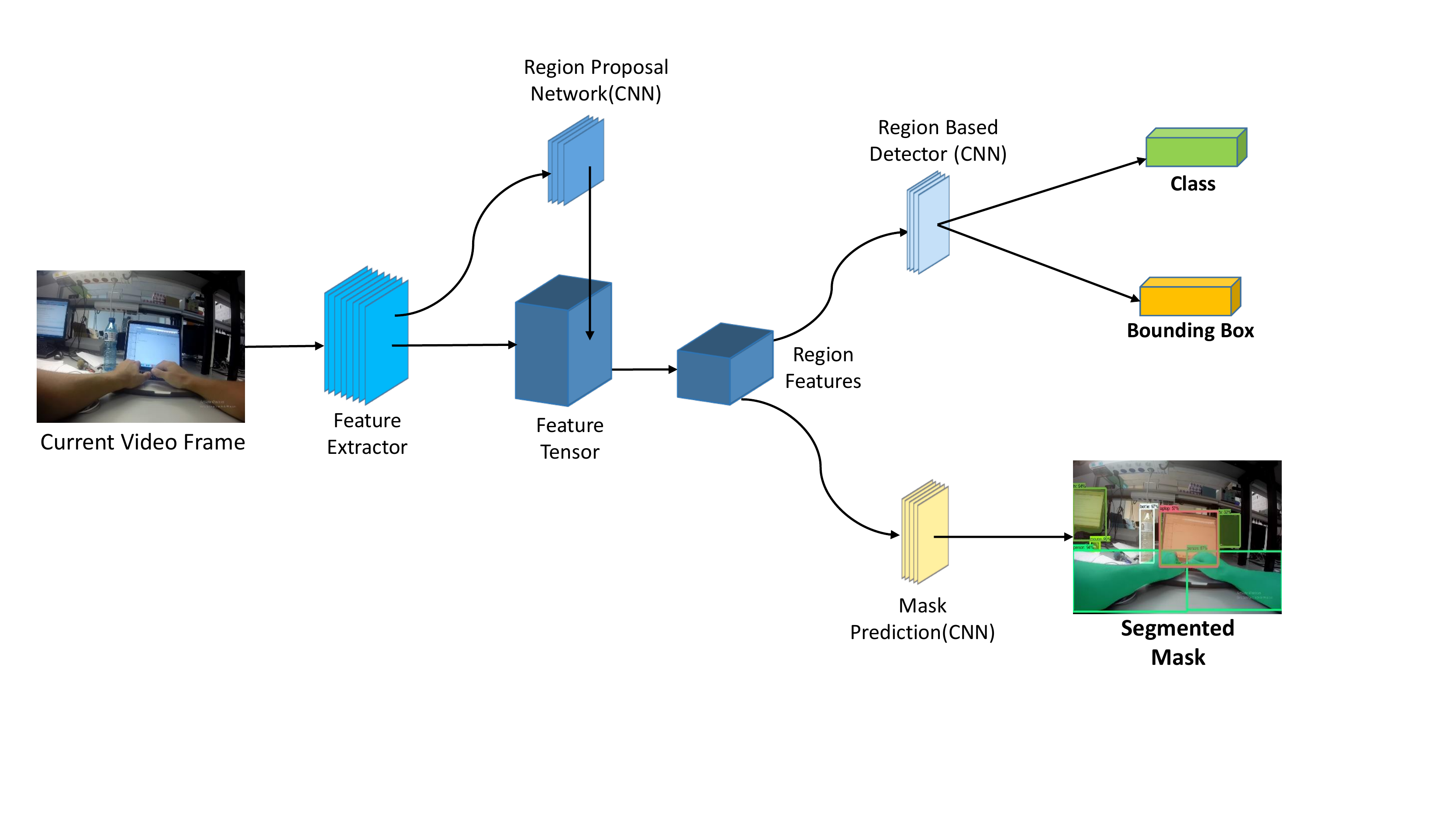}
    \caption{Model architecture for identification of sensitive regions from the video frames. A Mask R-CNN model with an Inception Resnet V2 Atrous feature extractor is used.}
    \label{sensitive_part_identification}
\end{figure}

The final architecture of the sensitive region segmentation is shown in Fig. \ref{sensitive_part_identification}. 
We used a pre-trained Inception-Resnet Hybrid based Mask R-CNN model trained on the COCO  dataset. First, we classify between 80 objects available in the COCO dataset. Next, we selected seven (7) potentially sensitive objects, as previously described. Finally, the frames are converted to RGB from BGR and are resized according to the required input dimension of the activity classifier network. 


\subsubsection{Protecting the sensitive objects}
\begin{figure}[t]
    \begin{subfigure}[t]{0.5\linewidth}
        \centering
        \includegraphics[height=1.2in,width=1.6in]{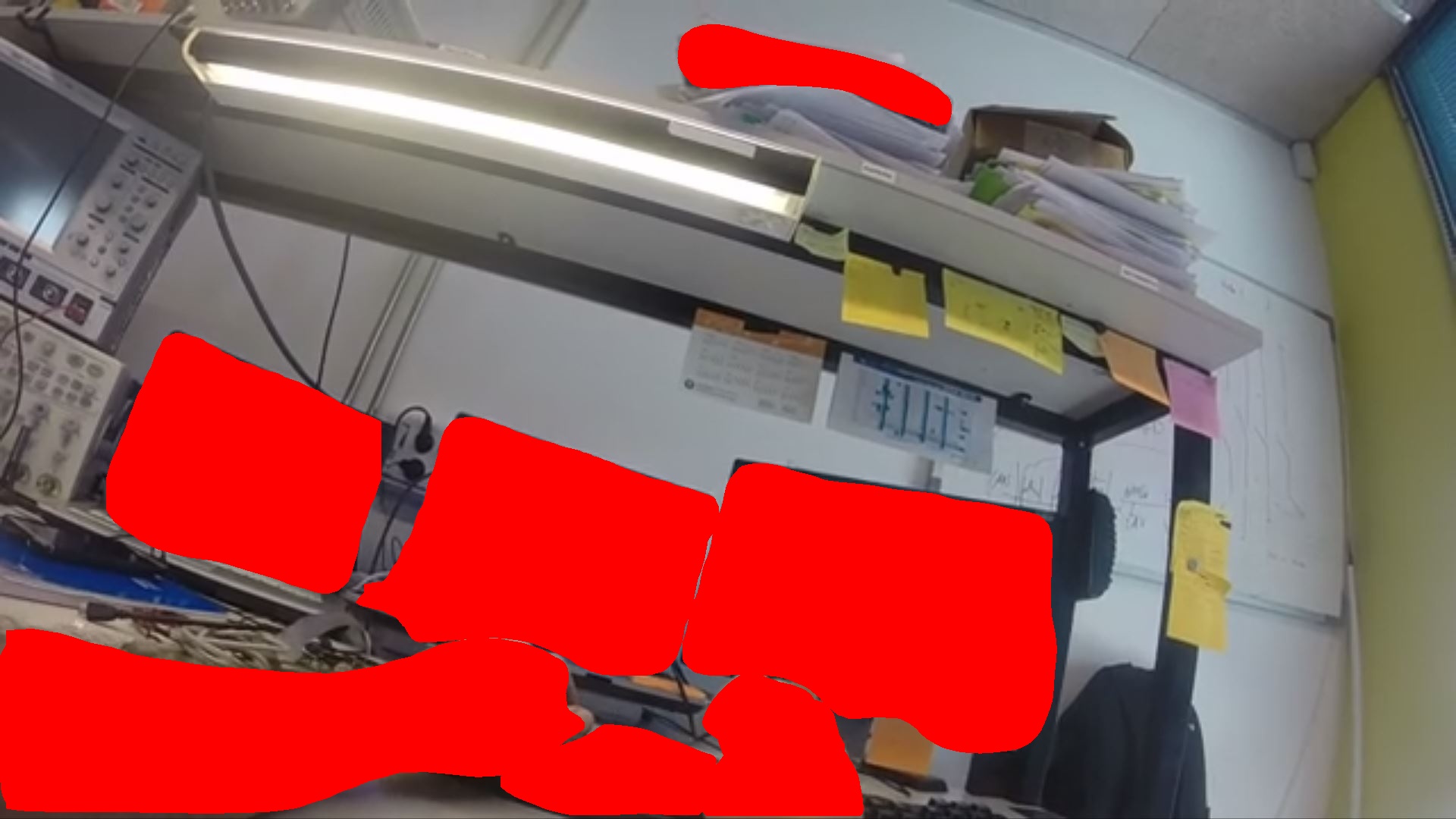}
        \caption{Identified mask}
    \end{subfigure}%
    \begin{subfigure}[t]{0.5\linewidth}
        \centering
        \includegraphics[height=1.2in,width=1.6in]{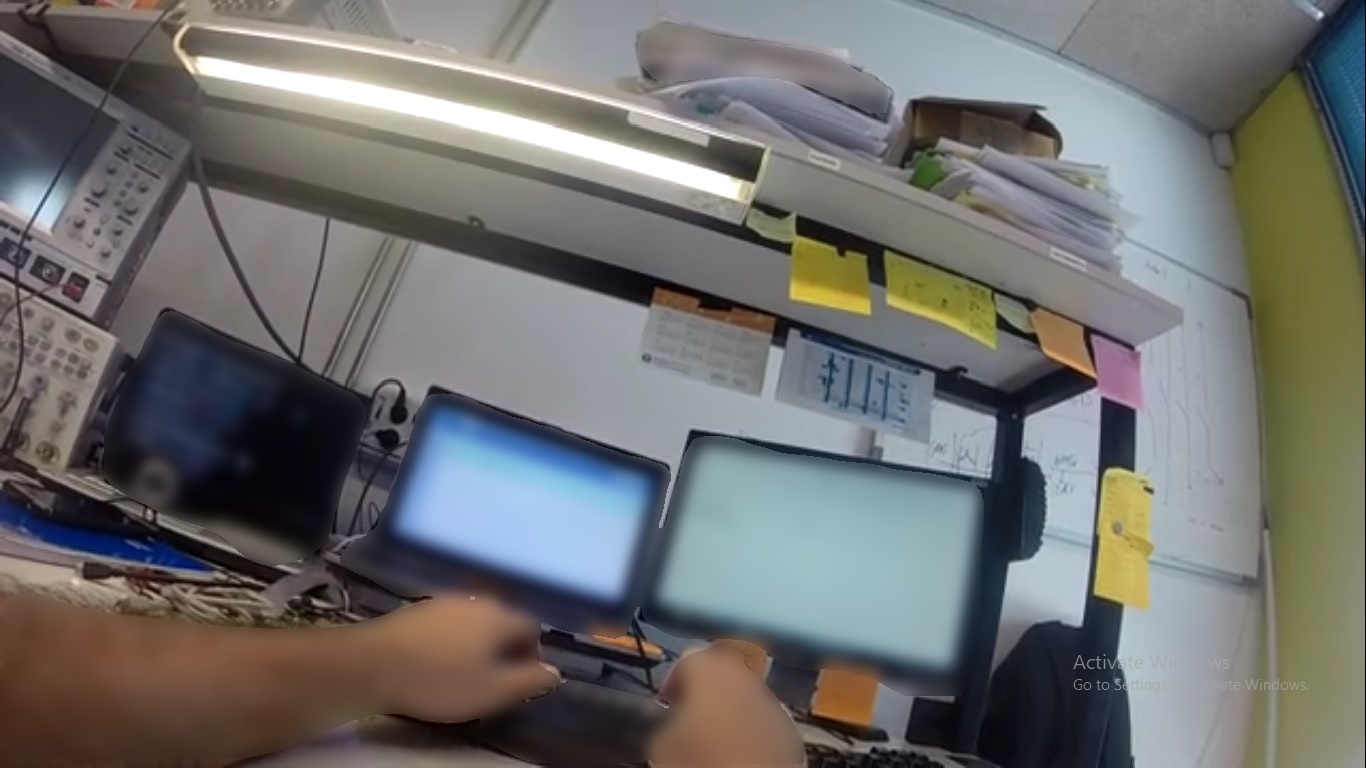}
        \caption{Sensitive regions blurred}
        \label{fig:InceptionResnetV2 Atrous}
    \end{subfigure}
    \caption{Example of generated masks using the Mask R-CNN network using the Inception ResNetV2 with Atrous convolution model as a feature extractor. The video frame was labeld as the ``Typeset" class.}
    \label{example_mask}
\end{figure}

After identifying the sensitive regions from the video frames, a Gaussian filter is used to blur the regions where the sensitive objects are detected. An example video frame from the ``chat" class is shown in Fig. \ref{example_mask} (a) along with the detected mask using the Mask R-CNN network, while Fig. \ref{example_mask} (b) shows the same frame with sensitive objects blurred (hands, screens, and keyboard). 

\subsection{Activity Classification Model}\label{activity_classification}
\begin{figure*}[t]
    \centering
    \includegraphics[width=0.9\linewidth]{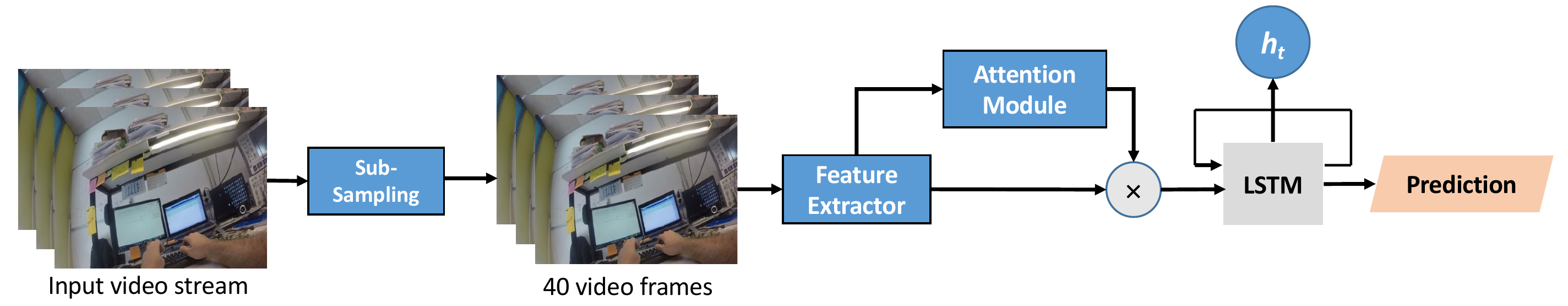}
    \caption{Proposed activity classification framework excluding the privacy protection module.}
    \label{network_activity_classification}
\end{figure*}

The proposed activity classification module consists of a 3-channel single stream network inspired by \cite{ullah2017action}. The network is illustrated in Fig. \ref{network_activity_classification}. The authors of \cite{ullah2017action} used AlexNet~\cite{krizhevsky2012imagenet} as the feature extractor with two uni-directional LSTM layers for temporal sequence modeling. 
In contrast, we propose to utilize an ensemble of Wide ResNet~\cite{zagoruyko2016wide}, ResNext~\cite{xie2017aggregated} and DenseNet~\cite{huang2017densely} feature extractors and a single layer bi-directional LSTM with framewise attention. 
The proposed framework is detailed in the following subsections.

\subsubsection{Video Frame Selection}
We select $40$ frames at random intervals from each video for training, as this setting provides the best overall results. These RGB image frames are then gamma-corrected and normalized before being loaded into the model.

\subsubsection{Feature Extraction Layer}
We experimented with different feature extractor models including, DenseNet121\cite{huang2017densely}, DenseNet161\cite{huang2017densely}, ResNet50\cite{he2016deep}, ResNet101\cite{he2016deep}, ResNet152\cite{he2016deep}, InceptionV3\cite{szegedy2016rethinking}, InceptionResNetV2\cite{szegedy2017inception}, ResNext\cite{xie2017aggregated}, Wide ResNet 101\cite{zagoruyko2016wide}. In these experiments, we used pre-trained models on ImageNet\cite{deng2009imagenet}. We froze the feature extractor and trained all the remaining layers of our model. 

\subsubsection{LSTM Layer}
We pass the video frames to a Bi-Directional single layer LSTM network\cite{huang2015bidirectional}. Our LSTM layer had an input dimension of $512$, as was returned by the feature extractor, and a hidden state size of $1024$. 

\subsubsection{Frame Wise Attention}
We hypothesize that, for activity classification, all video frames are not of equal importance. Accordingly, we use a frame-wise attention module with a sigmoid activation function after the LSTM layer.

\subsubsection{Fully Connected Layers}
The output of the frame-wise attention module is passed to the fully connected layer, followed by batch normalization and a softmax decision layer. 
\subsubsection{Training Scheme}
The proposed activity classification network is trained for $20$ epochs with the Adam optimizer \cite{kingma2014adam} using Cross-entropy as the loss function. 
To address the class-imbalance issue in the data, we implemented a balanced training scheme where every mini-batch contained an equal number of samples from each of the classes. 
We oversampled the underrepresented classes while creating these balanced mini-batches.
We begin training with an initial learning rate of $0.001$ and implemented a learning rate scheduler with a patience value of $5$ and with a decay factor of $1/10$ to facilitate model convergence. 




\section{Experimental Evaluation}

\subsection{Performance of Activity Classification Models}













In our activity classification network, we implemented a variety of feature extractors and compared their results in different settings. During our experiments, we observed that each model tested performs best at a specific frame size, and frame-wise attention does not improve the performance of all models. The top-performing models and their performance on the \emph{original}-test data is shown in Table \ref{ensacc}.


\begin{table}[h]
\centering
\caption{Performance comparison of different networks on the \emph{original}-test dataset}
\label{ensacc}
\begin{tabular}{lcc}
\hline
Model    & Resolution  & Accuracy (\%) \\ 
\hline
1. ResNext  & $248\times248$ & $75.92$ \\

2. DenseNet  & $512\times512$ & $74.87$ \\

3. Wide ResNet 101  & $324\times324$ & $79.84$ \\

4. Wide ResNet 101 + Attention  & $324\times324$ & $75.39$ \\
\hline
\end{tabular}
\end{table}

\begin{table}[tb]
\centering
\caption{Accuracy of the proposed model ensemble on different test sets while using different training sets}
\label{ensacc2}
\begin{tabular}{c|cc}
\hline
\multirow{2}{*}{Ensemble Training Data} & \multicolumn{2}{c}{Accuracy (\%)} \\ 
\cline{2-3}
& \emph{Original}-test  & \emph{Blurred}-test  \\ 
\hline
\emph{Original}-train dataset  & $85.07$  & $68.8$ \\
\emph{Mixed}-train dataset  & $82.72$ & $75.8$  \\
\hline
\end{tabular}
\end{table}

\subsection{Performance of Model Ensemble}
We propose to use an ensemble of the four classifiers described in the previous section. Ensemble weights are calculated based on the class-wise F1 score of each model.
We first compare the performance of the ensemble using models trained and tested on different data subsets. The results provided in Table \ref{ensacc2} show that the ensemble model trained only on the \emph{original}-training set performs poorly on the \emph{blurred}-test set. However, training on \emph{mixed} dataset improves the \emph{blurred}-test performance with a drop of accuracy in the \emph{original}-test condition. We address this trade-off by adding fine-tuned models in our final ensemble. 



\subsection{Model Tuning and Final Ensemble}\label{model_tuning}
To address the issue of performance degradation on the \emph{blurred} test, we select our models trained on the \emph{original} sub-dataset and use transfer learning to fine-tune them using the \emph{blurred} (privacy protected) set. We prepare the final ensemble using the four original models mentioned in Table \ref{ensacc} and the proposed fine-tuned versions.

\subsection{Results and Discussion}
\begin{figure}[tb]
\centering
\includegraphics[width=\linewidth]{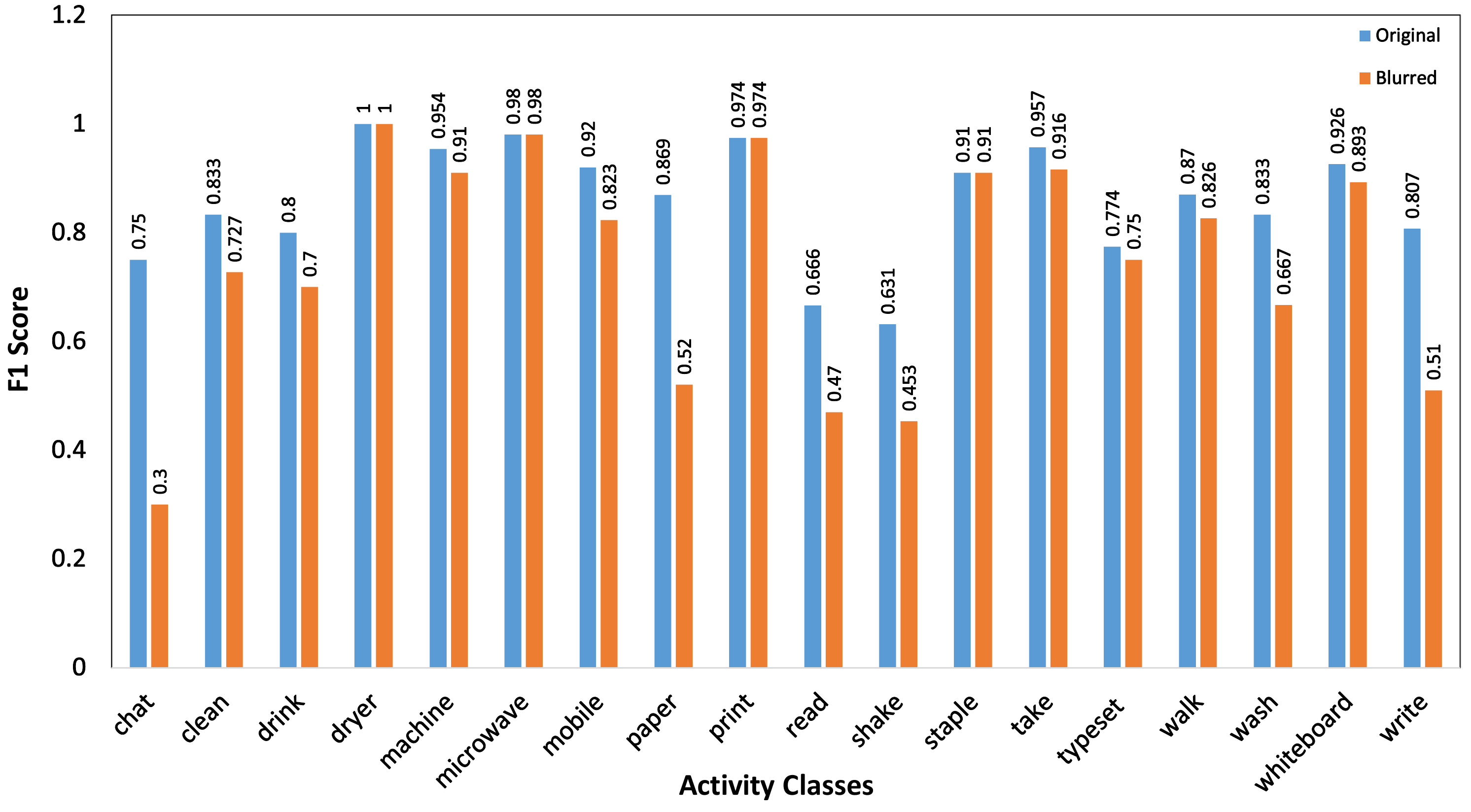}
\caption{F1 score distribution on \emph{original}-test and \emph{blurred}-test dataset videos across the class labels.}
\label{F1_distribution}
\end{figure}

The results of the final ensemble and class-wise F1 scores are shown in Table \ref{finalres} and Fig. \ref{F1_distribution}, respectively. From the results, we first observe that the class-imbalance did not affect our system performance. As an illustrative example, from Fig. \ref{class_piechart} we observe that the class ``typeset" included a significantly larger amount of data compared to the ``take" class. The final class-wise F1 scores depicted in Fig. \ref{F1_distribution} shows that the system performed better in detecting the minority class. Secondly, Fig. \ref{F1_distribution} shows that for most activity classes, the relative performance degradation due to privacy protection is not significant. On average, the performance metrics in Table \ref{finalres} degraded about $10\%$ from their original values due to privacy protection, demonstrating the effectiveness of the proposed privacy-aware activity classification system.



\begin{table}[tb]
\centering
\caption{Overall performance of the final ensemble classifier}
\label{finalres}
\begin{scriptsize}
\begin{tabular}{ccccc}
\hline
Sub-dataset & Precision & Recall  & F1 Score  & Accuracy (\%)\\ 
\hline
\emph{Original}  & $.88 (\pm0.1)$ & $.85 (\pm0.1)$ & $.86 (\pm0.1)$ & $85.08 (\pm11.4)$ \\
\emph{Blurred} & $.79 (\pm0.2)$ & $.75 (\pm0.2)$ & $.74 (\pm0.2)$ & $73.68 (\pm19.97)$ \\
\hline
\end{tabular}
\end{scriptsize}
\end{table}

\section{Conclusions}
In this work, we have developed a privacy-aware activity classification from FPV videos using an ensemble of deep learning models. The privacy protection module utilized pre-trained networks to identify sensitive regions from within the video frames and performed Gaussian blurring on those pixels. The system has been trained using balanced mini-batches to effectively address the issue of class imbalance in the training data. The ensemble of models trained on unprotected videos and later fine-tuned on privacy-protected videos provided the best overall performance in both conditions. 


\section{Acknowledgement}
We would like to thank the department of BME and Brain Station 23 (Dhaka, Bangladesh) for supporting this research. The TITAN Xp GPU used for this work was donated by the NVIDIA Corporation.

\balance

\bibliographystyle{IEEEtran}
\bibliography{ref.bib}
\balance 
\end{document}